\pdfoutput=1

\documentclass[11pt]{article}

\usepackage{emnlp2021}
\usepackage{bm}
\usepackage{times}
\usepackage{booktabs}
\usepackage{latexsym}
\usepackage{algorithm}
\usepackage{algorithmicx}  
\usepackage{algpseudocode}
\usepackage{multirow}
\usepackage{graphicx}
\usepackage{amsmath}
\usepackage{bbm}
\usepackage[T1]{fontenc}

\usepackage[utf8]{inputenc}

\usepackage{microtype}

\usepackage[normalem]{ulem}

\newcommand{\MODEL}{bert2BERT}

\title{\MODEL: Towards Reusable Pretrained Language Models}

\author{
 Cheng~Chen$^{1}\footnotemark[2]$, Yichun~Yin$^{2}$, Lifeng~Shang$^{2}$, Xin~Jiang$^{2}$, Yujia~Qin$^1$,  \\
 \textbf{Fengyu~Wang$^1$, Zhi~Wang$^3$, Xiao~Chen$^2$, Zhiyuan~Liu$^1$, Qun~Liu$^2$} \\
 $^1$Department of Computer Science and Technology, Tsinghua University \\
 $^2$Huawei Noah’s Ark Lab \\
 $^3$Tsinghua Shenzhen International Graduate School,Tsinghua University \\
 
}

\begin{document}
\maketitle
\begin{abstract}
In recent years, researchers tend to pre-train ever-larger language models to explore the upper limit of deep models. However, large language model pre-training costs intensive computational resources and most of the models are trained from scratch without reusing the existing pre-trained models, which is wasteful. In this paper, we propose \MODEL, which can effectively transfer the knowledge of an existing smaller pre-trained model (e.g., BERT$_{\rm BASE}$) to a large model (e.g., BERT$_{\rm LARGE}$) through parameter initialization and significantly improve the pre-training efficiency of the large model. Specifically, we extend the previous {\it function-preserving}~\cite{net2net} on Transformer-based language model, and further improve it by proposing {\it advanced knowledge} for large model's initialization. In addition, a two-stage pre-training method is proposed to further accelerate the training process. We did extensive experiments on representative PLMs (e.g., BERT and GPT) and demonstrate that (1) our method can save a significant amount of training cost compared with baselines including learning from scratch, StackBERT~\cite{stack} and MSLT~\cite{stack2}; (2) our method is generic and applicable to different types of pre-trained models. In particular, \MODEL\ saves about 45\% and 47\% computational cost of pre-training BERT$_{\rm BASE}$ and GPT$_{\rm BASE}$ by reusing the models of almost their half sizes. The source code will be publicly available upon publication.
\end{abstract}

\renewcommand{\thefootnote}{\fnsymbol{footnote}}
\footnotetext[2]{This work is done when Cheng Chen is an intern at Huawei Noah's Ark Lab}

\section{Introduction}

\begin{figure}[t]
\centering
	\includegraphics[width=0.44\textwidth]{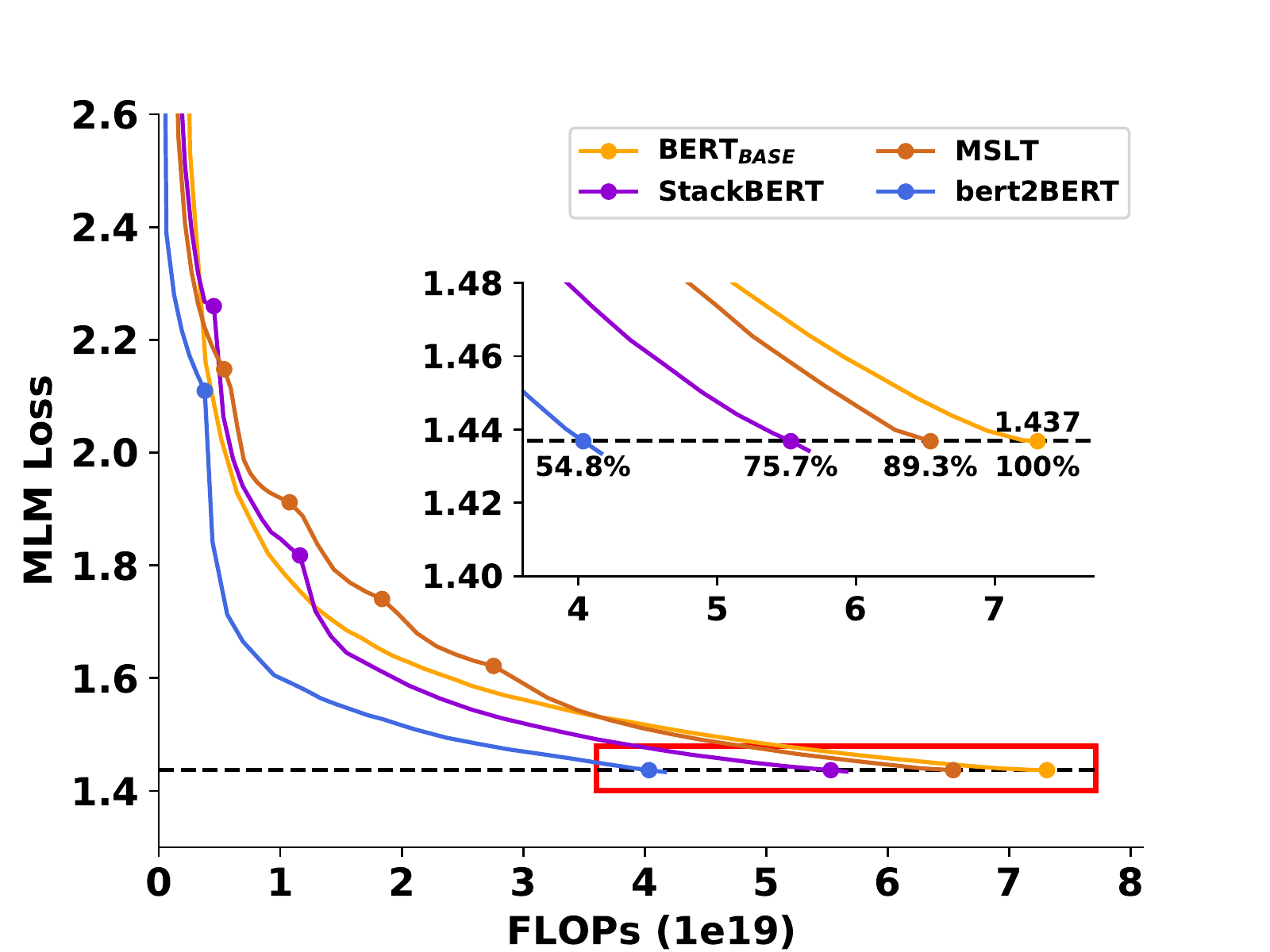}
	\caption{Loss curves of \MODEL\ and baselines. Both stackBERT~\citep{stack} and MSLT~\citep{stack2} are based on the progressive training setting. More details are shown in Table~\ref{tab:glue_results}.}
	\label{fig:pretrain_loss_curves_page1}
\end{figure}


\begin{figure*}[t]
	\includegraphics[width=\textwidth]{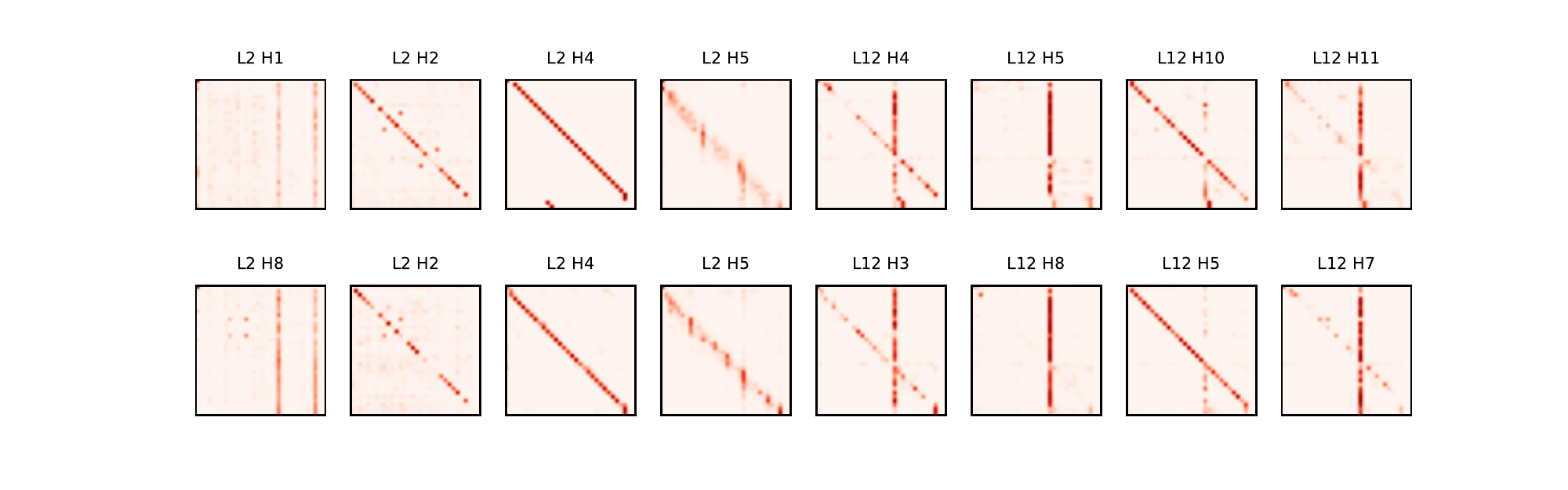}
	\caption{The comparisons of attention patterns between small and large PLMs. The upper ones are the attention patterns of BERT$_{\rm BASE}$ model of a architecture of \{$L$=$12$, $D$=$768$\}, and the lower ones are the attention patterns of one small BERT model whose architecture is \{$L$=$12$, $D$=$512$\}. We find that there are a large number of similar attention patterns in the same layer of the two models,
	 indicating the possibility of reusing parameters of trained small PLMs to speed up the pre-training of large PLMs.}
	\centering
	\label{fig:attention_maps_compare}
\end{figure*}

Pre-trained language models (PLMs), such as BERT~\citep{bert}, GPT~\citep{radford2018improving,gpt2,gpt3}, ELECTRA~\citep{electra}, XLNet~\citep{xlnet} and RoBERTa~\citep{roberta}, have achieved great success in natural language processing (NLP). However, the pre-training process of large PLMs can be extremely computationally expensive and produces huge carbon footprints. For example, GPT-3 was trained using 3.1E+6 GPU hours, at an estimated cost of \$4.6 million\footnote{\url{https://lambdalabs.com/blog/demystifying-gpt-3/}}, which will produce huge CO$_{2}$ emissions. 
Therefore, how to reduce the training cost of PLM is of great importance to Green AI~\citep{green}.


Recently, there is a trend of training extremely large models to explore the upper limits of PLMs. For example, large pre-trained models, including GPT-3~\citep{gpt3} (175B), PanGu-$\alpha$~\citep{pangu} (200B) and Switch Transformers~\citep{switch} (1571B), have been proved promising in language understanding and generation. However, these models are all pre-trained from scratch independently without utilizing the knowledge of smaller ones that have already been trained. On the other hand, our empirical studies show that the pre-trained models of different scales could share similar knowledge, for example in Figure~\ref{fig:attention_maps_compare}, the attention patterns of the two PLMs with different sizes are similar.

To save the training cost of large models, we propose the \MODEL\ method, which can efficiently transfer the learned knowledge of the smaller model to the large model. bert2BERT consists of two components: (1) for parameter initialization, we first extend the function preserving training~\citep{net2net} to PLMs by duplicating and stacking the parameters of the existing smaller PLM, which we call function-preserving initialization (FPI). FPI ensures that the initialized large model has almost the same behavior with the small model, so that the large model has a good starting point for later optimization. We also find that duplicating the weights of different layers can further accelerate the convergence of the large model, which we call advanced knowledge initialization (AKI). Although the strategy of AKI somewhat violates the principle of function preserving, we find that empirically it leads to a faster convergence rate and achieves higher training efficiency. (2) Secondly, a two-stage training strategy is further applied on the large model to accelerate the training process.


To demonstrate the superiority of our method, we conduct extensive experiments on two representative PLMs: BERT and GPT, with different model sizes. The results show that: (1) our method can save a significant amount of computation in pre-training compared to the traditional way of learning from scratch and progressive stacking methods such as StackBERT~\citep{stack} and MSLT~\citep{stack2}; (2) our method is model-agnostic, which can be applied on a wide range of Transformer-based PLMs. One typical example is that, when using a small pre-trained model with half the size of BERT$_{\rm BASE}$ for initialization, \MODEL\ saves {\bf 45\%} computation cost of the original BERT$_{\rm BASE}$ pre-training.

In general, our contributions are summarized as follows: (1) we explore a new direction for the efficient pre-training by reusing the trained parameters of small models to initialize the large model; (2) we successfully extend function preserving method~\citep{net2net} on BERT and further propose advanced knowledge initialization, which can effectively transfer the knowledge of the trained small model to the big model and improve the pre-training efficiency; (3) the proposed method outperforms other progressive training methods and achieves 45\% computation reduction on BERT$_{\rm BASE}$; (4) our method is generic, effective for both the BERT and GPT models, and have great potential to become an energy-efficient solution for pre-training super large-scale language models.

\section{Preliminary}
Before presenting our method, we first introduce some details about the BERT architecture, consisting of one embedding layer and multiple Transformer~\citep{attention} layers. 
\subsection{Embedding Layer} 
The embedding layer first maps the tokens in a sentence into vectors with an embedding matrix $\bm{W}^E$. Then one normalization layer is employed to produce the initial hidden states $\bm{H}_0$.

\subsection{Transformer Layer} 
The hidden states are iteratively processed by multiple Transformer layers as follows:
\begin{equation}
\setlength\abovedisplayskip{5pt}
\setlength\belowdisplayskip{5pt}
    \bm{H}_l = {\rm Transformer}_l(\bm{H}_{l-1}), l\in \left[1, L\right]
\end{equation}
where $L$ denotes the number of Transformer layers, each including a multi-head attention (MHA) and a feed-forward network (FFN).

\paragraph{MHA.} MHA is composed of multiple parallel self-attention heads. The hidden states of the previous layer are fed into each head and then the output of all heads is summed to obtain the final output as follows:
\begin{equation}\label{eq:MHA}
\setlength\abovedisplayskip{5pt}
\setlength\belowdisplayskip{5pt}
\begin{split}
   &\bm{Q}_i, \bm{K}_i, \bm{V}_i = {\bm{H}_{l-1}\bm{W}_{l,i}^Q, \bm{H}_{l-1}\bm{W}_{l,i}^K, \bm{H}_{l-1}\bm{W}_{l,i}^V }, \\
   &\bm{H}^{\rm HEAD}_{l,i} = {\rm softmax}(\frac{\bm{Q}_i {\bm{K}_i}^{T}}{\sqrt{d_k}})\bm{V}_{i} \bm{W}_{l,i}^O, \\
   &{\rm MHA}(\bm{H}_{l-1}) = \sum_{i=1}^a{\bm{H}^{\rm HEAD}_{l,i}}, \\
   &\bm{H}^{\rm MHA}_{l} = {\rm LayerNorm}(\bm{H}_{l-1} + {\rm MHA}(\bm{H}_{l-1})).
\end{split}
\end{equation}
The output of previous layer $\bm{H}_{l-1}$ is linearly projected to queries ($\bm{Q}_i$), keys ($\bm{K}_i$) and values ($\bm{V}_i$) using different weights $\bm{W}^Q_{l,i}, \bm{W}^K_{l,i}, \bm{W}^V_{l,i}$ respectively. $\bm{H}^{\rm HEAD}_{l,i}$ indicates the context-aware vector which is obtained by the scaled dot-product of queries and keys in the $i$-th attention head. $a$ represents the number of self-attention heads. $d_k$ is the head dimension acting as the scaling factor. $\bm{H}^{\rm MHA}_l$ is the input of the next sub-module FFN.
\vspace{2mm}

\paragraph{FFN.} FFN consists of two linear layers and one GeLU activation function~\citep{gelu}, which is formatted as: 
\begin{normalsize}
\begin{equation}\label{eq:FFN}
\setlength\abovedisplayskip{5pt}
\setlength\belowdisplayskip{5pt}
\begin{split}
    &\bm{H}^{\rm FFN}_{l} = {\rm GeLU}(\bm{H}^{\rm MHA}_{l}\bm{W}_l^1 + \bm{b}_{l}^1)\bm{W}_{l}^2 + \bm{b}_{l}^2,\\
    &\bm{H}_l = {\rm LayerNorm}(\bm{H}^{\rm MHA}_{l} + \bm{H}^{\rm FFN}_{l}).
\end{split}
\end{equation}
\end{normalsize}

\paragraph{Layer Normalization.} Both the modules of MHA and FFN have one layer normalization~\citep{ln} that stabilizes the hidden states dynamics in Transformer. Formally, it is written as:
\begin{equation}\label{eq:layernorm}
\setlength\abovedisplayskip{5pt}
\setlength\belowdisplayskip{5pt}
\begin{split}
    {\rm LayerNorm}(\bm{H}) &= (\frac{\bm{H}-\mu_{\bm H} }{\sigma_{\bm H} })\odot \bm{W}^{LN} + \bm{b}^{LN},
\end{split}
\end{equation}
where $\odot$ means the element-wise multiplication, the statistics of $\mu_{\bm H}$ and $\sigma_{\bm H}$ are the calculated mean and variance of hidden states $\bm{H}$ respectively.

\section{Methodology}

\subsection{Problem Statement}
We aim to accelerate the pre-training of target model $\mathcal{T}(L^t, D^t)$ by transferring the knowledge of an existing pre-trained model $\mathcal{S}( L^s, D^s)$, where $\ L^{s|t}$ means the numbers of Transformer layer and $D^{s|t}$ means the model width (i.e., hidden size), satisfying $L^s \leq L^t$ and $D^s \leq D^t$. Formally, our problem are two-fold: (1) how to perform an effective parameter initialization for $\mathcal{T}$ by reusing the trained parameters of $\mathcal{S}$, and (2) how to efficiently train the initialized $\mathcal{T}$, so that $\mathcal{T}$ can have a faster convergence rate in pre-training.

\subsection{Overview}

Targeting the above problems, \MODEL\ first initializes the target model $\mathcal{T}$ with the parameters of the existing model $\mathcal{S}$ by the width-wise expansion ($D^s \rightarrow D^t$) and depth-wise expansion ($L^s\rightarrow L^t$). Through this expansion, the knowledge contained in the parameters of the source model is directly transferred to the target model. Then we further pre-train the initialized target model with a two-stage pre-training method. We illustrate the overall workflow in section~\ref{sec:efficient_train}.

\begin{figure}[t]
	\centering
	\includegraphics[width=0.45\textwidth]{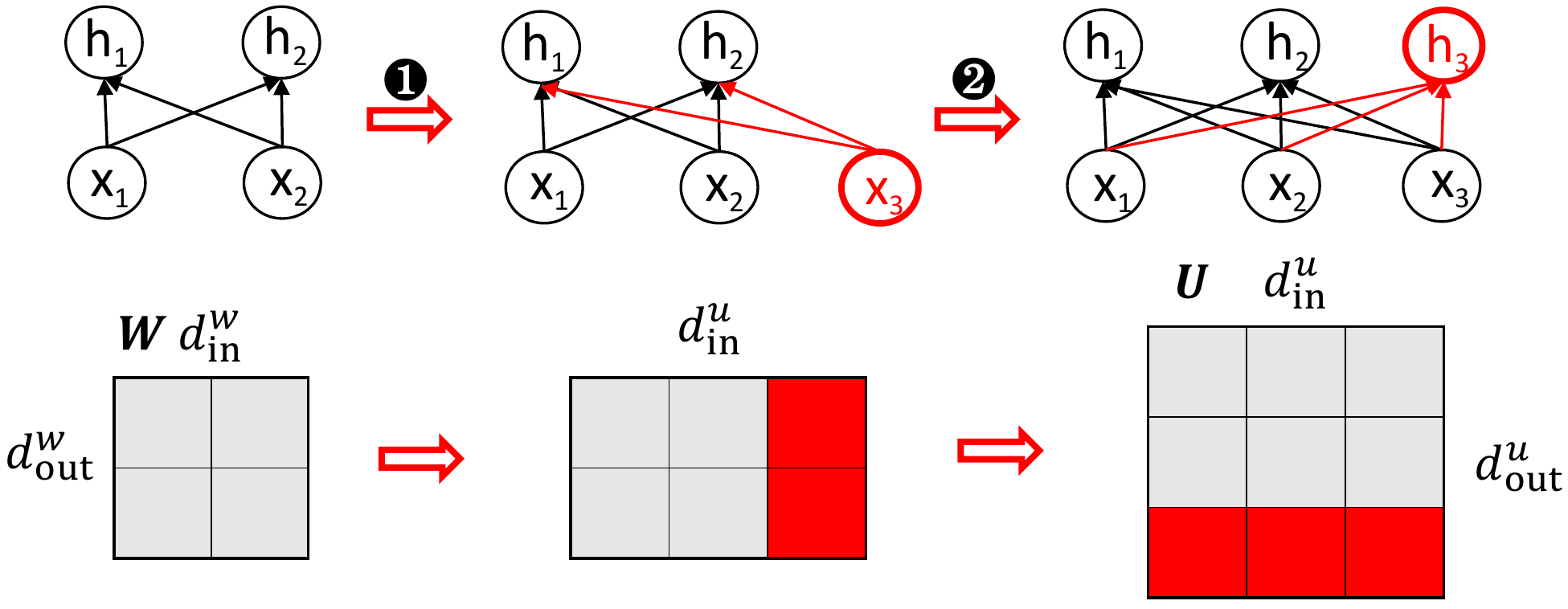}
	\caption{Overview of the matrix expansion. It first enlarges the size of in-dimension and then enlarges the size of out-dimension. From the view of neurons, it adds new neurons (red) in both input and output.}
	\label{fig:matrix_expansion_0}
\end{figure}

Essentially, the width-wise expansion can be decomposed into expansions of a set of matrices (or vectors\footnote{We omit the expansion of bias (vector) for simplicity. It follows a similar process as the matrix expansion.}). As illustrated in Figure~\ref{fig:matrix_expansion_0}, the matrix expansion enlarges a parameter matrix ${\bm W}\in \mathbbm{R}^{d_{\rm in}^w*d_{\rm out}^w}$ of $\mathcal{S}$ to ${\bm U}\in \mathbbm{R}^{d_{\rm in}^u*d_{\rm out}^u}$ of $\mathcal{T}$ by two kinds of operations: in-dimension and out-dimension expansion.

In the following sections, we first introduce two strategies of width-wise expansion: function-preserving and advanced knowledge initialization. Then, we introduce the depth-wise expansion and detail the two-stage pre-training process.





\subsection{\MODEL}
Before presenting the proposed method, we introduce two index mapping functions: $g_{\rm in}$ and $g_{\rm out}$, where $g_{\rm in}(i)$ means the $i$-th column of ${\bm U}$ reuses the $g_{\rm in}(i)$-th column parameters of ${\bm W}$, $g_{\rm out}(j)$ means the $j$-th row of ${\bm U}$ reuses the $g_{\rm out}(j)$-th row parameters of ${\bm W}$. Both methods are defined with these two mapping functions.

\begin{figure}[t]
\centering
	\includegraphics[width=0.45\textwidth]{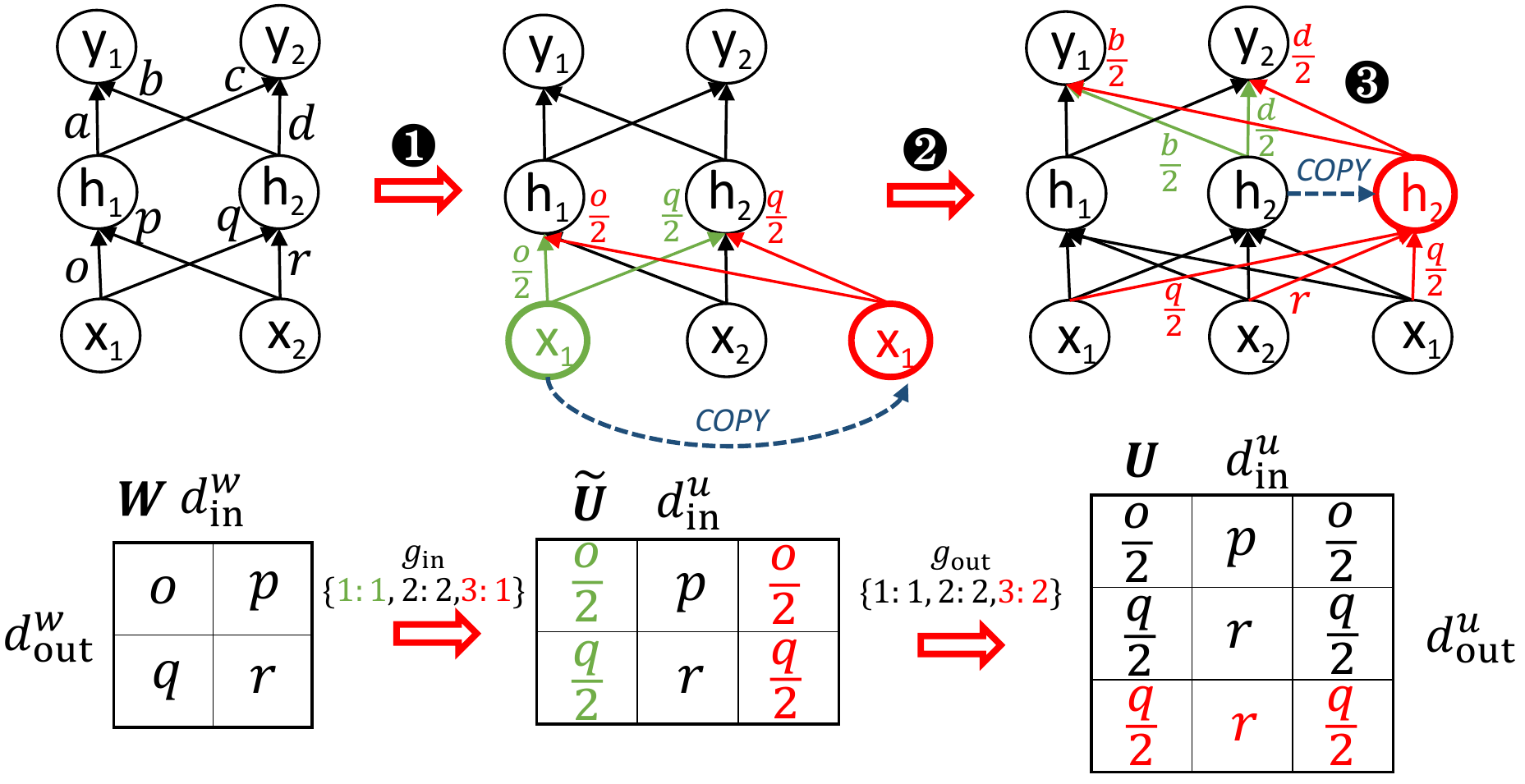}
	\caption{Overview of the function preserving initialization (FPI). Given the same input \{$x_1$, $x_2$\}, FPI ensures the initialized target model has the same output of \{$y_1$, $y_2$\} with the source model. The first and the second steps are expanding the in-dimension and out-dimension of the parameter matrix according to both $g_{in}$ and $g_{out}$ mapping functions respectively. From the view of neurons, FPI copies the input and output neurons to expand the network. After the matrix expansion, we can further use the in-dimension expansion to ensure the output same as the original one.}
	\centering
	\label{fig:matrix_expansion_1}
\end{figure}

\subsubsection{Function Preserving Initialization}

Function preserving initialization (FPI)~\citep{net2net} aims to make the initialized target model have the same function as the source model, which means that given the same input, the initialized target model has the same output as the source model. 
We give an example in Figure~\ref{fig:matrix_expansion_1} to illustrate FPI. Formally, the mapping functions are defined as follows: 
\begin{equation}
\setlength\abovedisplayskip{5pt}
\setlength\belowdisplayskip{5pt}
g_{\rm in}(i)=
\begin{cases}
i& i\in [1, d_{\rm in}^w]\\
f(\{1, 2,...,d_{\rm in}^w\}) & i\in (d_{\rm in}^w, d_{\rm in}^u],
\end{cases} 
\end{equation}
\begin{equation}
\setlength\abovedisplayskip{5pt}
\setlength\belowdisplayskip{5pt}
g_{\rm out}(j)=
\begin{cases}
j& j\in[1, d_{\rm out}^w]\\
f(\{1, 2,...,d_{\rm out}^w\}) & j\in (d_{\rm out}^w, d_{\rm out}^u],
\end{cases}
\end{equation}
where $f(\cdot)$ means the uniform sampling. We hereby denote the weight expansion as
$$\bm{U}={\rm EXPN}(\bm{W}; g_{\rm in}, g_{\rm out}),$$ which includes two stages of expansion, in-dimension expansion \eqref{eq:fpi_in} and out-dimension expansion \eqref{eq:fpi_out}:
\begin{equation}\label{eq:fpi_in}
\setlength\abovedisplayskip{5pt}
\setlength\belowdisplayskip{5pt}
\begin{split}
C_{g_{\rm in}(i)} &= \sum_{i'=1}^{d_{\rm in}^u}\mathbbm{I}({g_{\rm in}(i')= g_{\rm in}(i)}) \\
\bm{\widetilde U}_{(i, *)}&=\frac{1}{C_{g_{\rm in}(i)}}\bm{W}_{(g_{\rm in}(i), *)}, \\
\end{split}
\end{equation}
\begin{equation}\label{eq:fpi_out}
\bm{U}_{(*, j)}=\bm{\widetilde U}_{(*, g_{\rm out}(j))},
\end{equation}
where $\mathbbm{I}(\cdot)$ is an indicator function, and $C_{g_{\rm in}(i)}$ is the count of $g_{\rm in}(i)$ in the values of $g_{\rm in}$, which is used to re-scale the original parameters for keeping the function preserving property.

\paragraph{Expansion for all modules.} We apply FPI for all modules of BERT via matrix expansion ${\rm EXPN(\cdot)}$. Specifically, for the embedding matrix ${\bm{U}^E} $, we only conduct the out-dimension expansion:  
\begin{equation}~\label{eq:emb_fpi}
\setlength\abovedisplayskip{5pt}
\setlength\belowdisplayskip{5pt}
    \bm{U}_{(*, j)}^E=\bm{W}_{(*, g_{\rm out}^e(j))}^E.
\end{equation}

MHA module can be decomposed into multiple parallel self-attention heads and we conduct the head expansion for this module. The head expansion is formulated as: 
\begin{equation}
\setlength\abovedisplayskip{5pt}
\setlength\belowdisplayskip{5pt}
\begin{split}
\bm{U}^{Q|K|V}&={\rm EXPN}(\bm{W}^{Q|K|V}; g_{\rm in}^{q|k|v}, g_{\rm out}^{q|k|v}), \\
\bm{U}^{O}&={\rm EXPN}(\bm{W}^O; g_{\rm in}^o, g_{\rm out}^o). \\
\end{split}
\end{equation}

MHA module performs the head-wise expansion which means that we reuse the head parameters ($\bm{Q}_i$|$\bm{K}_i$|$\bm{V}_i$ in Eq.~\ref{eq:MHA}) to construct the matrices of new heads. The out-dimension expansion is:
\begin{equation}
\setlength\abovedisplayskip{5pt}
\setlength\belowdisplayskip{5pt}
g^{q|k|v}_{\rm out}(j)=
\begin{cases}
j& j\in[1, a^s]\\
f(\{1, 2,...,a^s\}) & j\in (a^s, a^t],
\end{cases}
\end{equation}
where $a^{s|t}$ mean that head numbers of source model and target model respectively. The module has three constraints: \{ $g_{\rm out}^e = g_{\rm in}^{q|k|v}$ ; $g_{\rm out}^{q|k|v} = g_{\rm in}^o$; $g_{\rm in}^{q|k|v} = g_{\rm out}^o$\}, with the first two constraints for the dimension consistency and the third one for residual connection.

For the FFN module, we perform the expansion on the parameter matrices ${\bm W}^{1|2}$ (Eq.~\ref{eq:FFN}) as follows:
\begin{equation}
\setlength\abovedisplayskip{5pt}
\setlength\belowdisplayskip{5pt}
\begin{split}
\bm{U}^{1}&={\rm EXPN}(\bm{W}^1; g_{\rm in}^1, g_{\rm out}^1), \\
\bm{U}^{2}&={\rm EXPN}(\bm{W}^2; g_{\rm in}^2, g_{\rm out}^2). \\
\end{split}
\end{equation}
Similar to the MHA module, the mapping functions of FFN also have three constraints: \{$g_{\rm out}^o = g_{\rm in}^1$; $g_{\rm out}^1 = g_{\rm in}^2$; $g_{\rm in}^1 = g_{\rm out}^2$\}. 

Layer normalization is a common module stacked on the top of MHA and FFN. Taking the layer normalization of FFN as an example, its expansion is formulated as:
\begin{equation}
\setlength\abovedisplayskip{5pt}
\setlength\belowdisplayskip{5pt}
\begin{split}
\bm{U}^{\rm ln}_{j}&=\bm{W}^{\rm ln}_{g^2_{\rm out}(j)}.\\
\end{split}
\end{equation}
Note that in layer normalization (Eq.~\ref{eq:layernorm}), we need to re-calculate the mean $\mu$ and variance $\sigma$ of hidden representations $\bm H$. The expansion of this parameter inevitably induces a gap and prevents the target model from strictly following the function preserving principle. However, we empirically find that the gap is so small that it can hardly affect the initialization and convergence of the target model. Thus we ignore this discrepancy.

We have validated the effectiveness of the proposed FPI in different settings in Table~\ref{tab:initial_loss}. The results show that the initialized model $\mathcal{T}$ achieves almost the same loss as $\mathcal{S}$, demonstrating that FPI successfully retains the knowledge of the small model when performing parameter expansion.

\begin{table}[t]
\centering
\resizebox{0.32\textwidth}{!}{
\begin{tabular}{l|c|c}
\toprule
Method & $\mathcal{S}(12, 384)$  & $\mathcal{S}(12, 512)$ \\
\midrule
Original &1.89 &1.67\\
\midrule
Rand &10.40 &10.42\\
DirectCopy  &9.05 &6.45\\
{\bf FPI} & {\bf 1.89} &{\bf 1.70}\\
\bottomrule
\end{tabular}}
\caption{The comparison of MLM losses between FPI and baselines. ``Original'' refers to the MLM losses of source pre-trained models $\mathcal{S}$. ``Rand'' refers to the MLM losses of randomly initialized target models.  ``DirectCopy'' refers to a naive method that directly copies the source model to the target model and the unfilled part is randomly initialized, ``FPI'' represents our function preserving method. We expand both models to the target model $\mathcal{T}(12, 768)$ and find that FPI can basically make the target model have similar losses with these trained source models. The loss gap between FPI and Original is brought by layer normalization.}
\label{tab:initial_loss}
\end{table}

\subsubsection{Advanced Knowledge Initialization}
\begin{figure}[t]
	\includegraphics[width=0.47\textwidth]{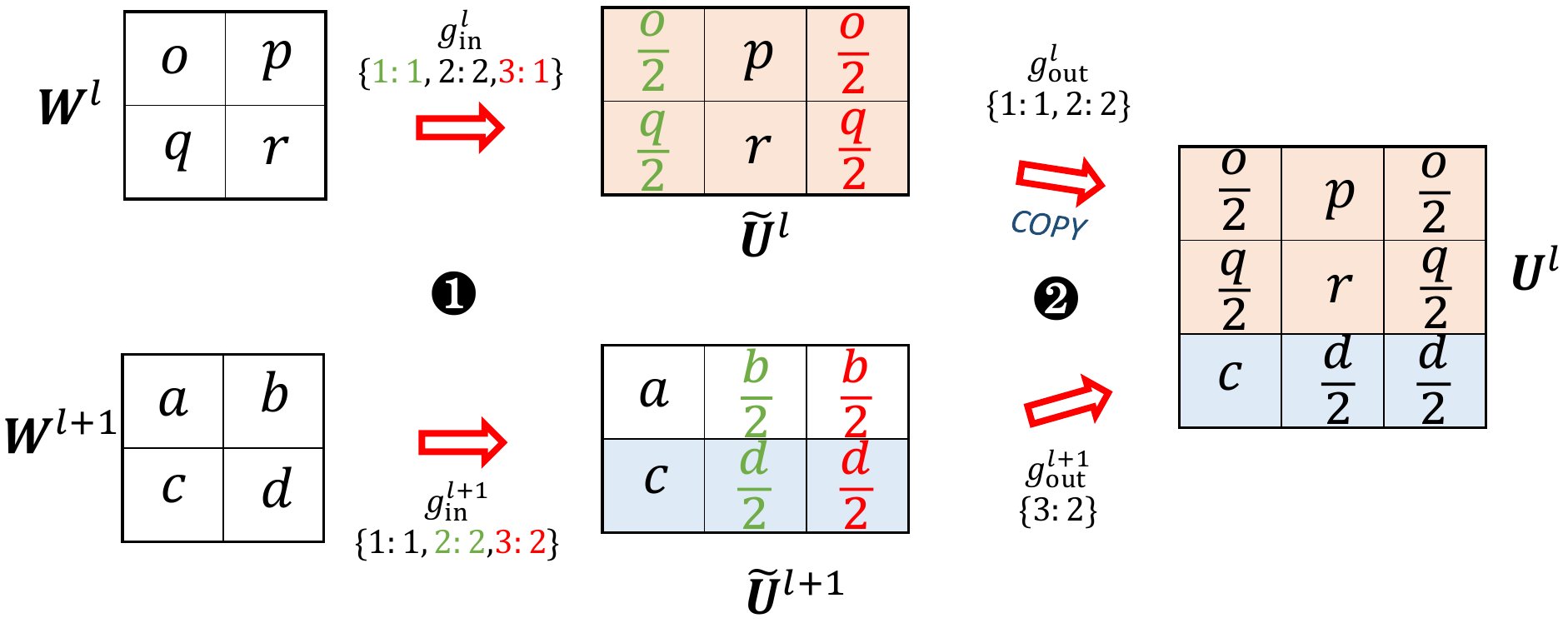}
	\caption{Overview of the advanced knowledge initialization (AKI). It first performs the in-dimension expansion on both the matrixes of current and upper layers. Then it uses the widened matrix of the current layer as the top part of the new matrix and samples the row of the widened matrix of the upper layer as the bottom part of the new matrix.}
	\centering
	\label{fig:matrix_expansion_2}
\end{figure}

To further improve the convergence rate of the pre-training target model, we propose the advanced knowledge initialization (AKI), which expands new matrices based on not only the parameters of the same layer but also the parameters of the upper layer in the source model. The intuition behind is based on previous findings~\citep{whatbertlearn, whatbertlook} that adjacent Transformer layers have similar functionality, which ensures that it will not damage the knowledge contained in the parameters of this layer. What's more, the knowledge comes from adjacent layers can break the symmetry~\citep{net2net} that FPI brings, which has been demonstrated beneficial. Specifically, we give an illustrative example in Figure~\ref{fig:matrix_expansion_2}.  
We formulate AKI as:
\begin{equation}
    \bm{U}^l={\rm EXPN}(\bm{W}^l, \bm{W}^{l+1}; g_{\rm in}, g_{\rm out}).
\end{equation}
Specifically, we first do the in-dimension expansion for both $\bm{W}^l$ and $\bm{W}^{l+1}$. Here we take $\bm{W}^l$ as an example:
\begin{equation}\label{eq:aki_1}
\setlength\abovedisplayskip{5pt}
\setlength\belowdisplayskip{5pt}
\begin{split}
C_{g_{\rm in}^l(i)} &= \sum_{i'=1}^{d_{\rm in}^u}\mathbbm{I}({g_{\rm in}^l(i')= g_{\rm in}^l(i)}) \\
\bm{\widetilde U}^l_{(i, *)}&=\frac{1}{C_{g_{\rm in}^l(i)}}\bm{W}^l_{(g_{\rm in}^l(i), *)}. \\
\end{split}
\end{equation}
Then we stack the expanded matrixes of $\bm{\widetilde U}^{l|l+1}$ to construct the final matrix:
\begin{equation}\label{eq:aki_2}
\setlength\abovedisplayskip{5pt}
\setlength\belowdisplayskip{5pt}
\begin{split}
\bm{U}^l_{(*, j)} &=
\begin{cases}
\bm{\widetilde U}^l_{(*, j)} & j\in [1, d_{\rm out}^s]\\
\bm{\widetilde U}^{l+1}_{(*, g_{\rm out}^{l+1}(j))} & j\in (d_{\rm out}^s, d_{\rm out}^t],
\end{cases}  \\
\end{split}
\end{equation}
Here we directly copy the whole expanded ${\bm{\widetilde U}^l}$ as the top part of the new matrix and place the sampled parameters from ${\bm{\widetilde U}^{l+1}}$ on the bottom of the new matrix.

We aggregate upper-layer information into a new matrix for two intuitions: (1) it breaks the FPI symmetry that hinders model convergence~\citep{net2net}; (2) upper-layer information can be used as high-level knowledge to guide the model to converge faster.
Empirically, we find that the AKI method outperforms FPI, while the performance is worse if we build a new matrix based on the matrix of the lower layer (or low-level knowledge). How to construct the optimal initialization for the target model with the parameters of different layers remains an open question and we leave it as future work.

\paragraph{Expansion for all modules.} For embedding matrix, we only do the out-dimension expansion as Eq.~\ref{eq:emb_fpi} in the FPI. Both the modules of MHA and FFN do the matrix expansion by following the defined operation Eq.~\ref{eq:aki_1}. and~\ref{eq:aki_2}. The constraints of mapping functions follow the setting of FPI.  

We display the attention patterns of the target model initialized by AKI in Appendix D and find that the target model can maintain the attention patterns of both current and upper layers very well.

\renewcommand{\algorithmicrequire}{\textbf{Input:}} 
\renewcommand{\algorithmicensure}{\textbf{Output:}} 

\subsubsection{Depth-wise Expansion}
After the width-wise expansion, we obtain a widened model with the same width as the target model. To bridge the depth gap, we perform depth-wise expansion to increase model depth to the depth of target model. We illustrate this process in Algorithm~\ref{alg:bert2bert} and the main idea is to iteratively stack the widened model until its depth is equal to the target model~\citep{stack}.



\subsubsection{Two-stage Pre-training} \label{sec:efficient_train}

\begin{algorithm}[t] 
  \caption{\MODEL\ initialization} 
	 \label{alg:bert2bert}
  \begin{algorithmic}[1] 
   \Require the target model $\mathcal{T}(L^t, D^t)$ and the source model $\mathcal{S}(L^s, D^s)$.
   \State $\mathcal{T}_1(L^s, D^t) \leftarrow$ do AKI or FPI with $\mathcal{S}( L^s, D^s)$
   \State $k \leftarrow \lfloor{L^t/L^s}\rfloor $
\For{$t = 2 \to k$}
	\State $\mathcal{T}_t(L^s\cdot t, D^t) \leftarrow $ stack $\mathcal{T}_1$ on top of $\mathcal{T}_{t-1}$
	\EndFor
	\State $\mathcal{T} \leftarrow$ stack top $L^t- L^s\cdot k $ layers of $\mathcal{T}_1$. 
  	\Ensure the initialized model $\mathcal{T}(L^t, D^t)$
  \end{algorithmic} 
 \end{algorithm}

To further improve the pre-training efficiency of initialized target model, we propose a two-stage training method: (1) We first train sub-structures of target model in a random manner to make the complete model converge at low cost. These substructures are built with bottom Transformer layers of target model and share one classification head. At each optimization step, we randomly sample one substructure and only update its top Transformer layers and the shared classification head. (2) After the sub-structure training, we further perform the traditional full-model training. The details of our method are displayed in Algorithm~\ref{alg:block_train}.


\begin{algorithm}[t] 
  \caption{Two-stage Pre-training} 
	 \label{alg:block_train}
  \begin{algorithmic}[1] 
   \Require the initialized model $\mathcal{T}$, large-scale unsupervised dataset $\mathcal{D}$, the epoch number of sub-model training $E_b$ and the epoch number of whole training process $E$, the layer number $l_b$. 
   \State Construct sub-models and these models have the layer numbers of \{$l_b, 2\cdot l_b,\dots, L^t$\}.
    \For{$e = 1 \to E_b$}
    		\For{$batch$ in $D$} 
    			\State $\mathcal{T}' \leftarrow$ sample one sub-model.
    			\State Perform forward and backward of $\mathcal{T}'$.
    			\State Update only top $l_b$ layers of $\mathcal{T}'$.
    		\EndFor
    	\EndFor
     \For{$e = E_b \to E$}
		\For{$batch$ in $D$}
			\State Perform forward and backward of $\mathcal{T}$.
			\State Update whole model $\mathcal{T}$.
		\EndFor
	\EndFor
  	\Ensure the pre-trained model $\mathcal{T}$
  \end{algorithmic} 
 \end{algorithm}

\section{Experiment}\label{sec:exp}

\begin{table*}
	\centering
	\scalebox{0.64}{
		\begin{tabular}{l|c c|cccccccc|c}
			\toprule
			Model & FLOPs & Ratio & SQuADv1.1 & SST-2 & MNLI & MRPC & CoLA & QNLI & QQP & STS-B & Avg. \\ 
			~ & ($\times$ ${\rm 1e19}$) & (Cost Saving) & (F1) & (Acc)& (Acc)& (Acc)& (Mcc)& (Acc)& (Acc)& (Acc)& ~ \\
			\midrule 
BERT$_{\rm BASE}$~(Google) & - & - & 88.4(0.1) & 93.6(0.2) & 84.7(0.1) & 87.9(0.9) & 59.6(1.5) & 91.6(0.1) & 91.4(0.1) & 89.6(0.5) & 85.8(0.1)  \\ 
BERT$_{\rm BASE}$ $\dag$~(Ours) & 7.3 & 0\% & 89.6(0.1) & 92.7(0.2) & 84.6(0.2) & 88.6(0.5) & 57.3(4.0) & 90.6(0.7) & 90.6(0.1) & 89.9(0.3) & 85.5(0.5)  \\ 
\midrule
\multicolumn{12}{l}{\it Progressive Training} \\
\midrule
{MSLT}\dag & 6.5 & 10.7\% & 90.4(0.2) & 92.9(0.2) & 85.1(0.2) & 87.9(2.1) & 55.6(4.1) & 90.7(0.2) & 90.6(0.2) & 88.2(0.6) & 85.2(0.7) \\
{StackBERT}\dag   & 5.5 & 24.3\% & 90.4(0.2) & 92.6(0.4) & 85.3(0.1) & 88.2(1.0) & 63.2(0.9) & 91.0(0.4) & 91.0(0.1) & 86.7(0.7) & 86.0(0.2) \\
\midrule
 \multicolumn{12}{l}{ {\it \MODEL\ }: $\mathcal{S}$(12, 512) $\rightarrow$ $\mathcal{T}$(12, 768)} \\
\midrule
DirectCopy & 6.4 & 12.2\% & 89.8(0.2) & 92.9(0.3) & 84.7(0.2) & 86.2(0.6) & 62.2(0.7) & 90.2(0.6) & 90.4(0.1) & 89.2(0.1) & 85.7(0.1) \\
${\bf FPI}$ & 5.1  & 30.4\% & 90.0(0.2) & 92.6(0.4) & 85.2(0.1) & 87.1(0.5) & 61.5(0.9) & 90.9(0.6) & 90.8(0.2) & 89.7(0.2) & 86.0(0.1) \\
${\bf AKI}$ & 4.5 & 38.4\% & 90.4(0.1) & 92.5(0.4) & 85.3(0.4) & 87.8(0.9) & 61.0(1.4) & 91.2(0.2) & 90.5(0.1) & 89.5(0.2) & 86.0(0.2) \\
${\bf \MODEL\ }$ &  $\bm{4.0}$  & $\bm{45.2\%}$ & 90.0(0.2) & 92.9(0.1) & 85.1(0.1) & 87.7(0.7) & 60.0(1.2) & 90.5(0.8) & 90.4(0.1) & 89.2(0.2) & 85.7(0.4) \\
\bottomrule
\end{tabular}}
\caption{Comparison between \MODEL\ and baselines. We report mean (and standard deviation) performance over 3 runs on the dev set. \MODEL\ means the combination of AKI and two-stage pre-training. FPI and AKI mean that the function preserving initialization, advanced knowledge initialization respectively. \dag~means the re-implemented results, where the BERT$_{\rm BASE}$ and StackBERT achieve similar results with the original paper, and MSLT result is different from the original paper because it uses a different optimizer (LAMB~\citep{lamb}) and only trains the corpus with a max sequence length of 128.}
\label{tab:glue_results}
\end{table*}

\subsection{Experimental Setup}
\paragraph{Pre-training Details.} We use the concatenation of English Wikipedia and Toronto Book Corpus~\citep{book} as the pre-training data. The settings of pretraining are: peak learning rate of 1e-4, warmup step of 10k, training epochs of $E$=40, batch size of 512, sub-model training epochs of $E_b$=5, layer number of $l_b$=3. Unless otherwise noted, all methods including bert2BERT and baselines use the same pre-training settings for fair comparisons. In the settings of bert2BERT, the target model has a BERT$_{\rm BASE}$ architecture of $\mathcal{T}(12, 768)$ and two architectures of source models of $\mathcal{S}(12, 512)$ and $\mathcal{S}(6, 512)$ are evaluated.

\paragraph{Fine-tuning Details.} For the evaluation, we use tasks from GLUE benchmark~\citep{wang2019glue} and SQuADv1.1~\citep{2016squad}. We report F1 for SQuADv1.1, Matthews correlation (Mcc) for CoLA~\citep{cola} and accuracy (Acc) for other tasks. For the GLUE fine-tuning, we set batch size to 32, choose the learning rate from \{5e-6, 1e-5, 2e-5, 3e-5\} and epochs from \{4, 5, 10\}. For the SQuADv1.1 fine-tuning, we set the batch size to 16, the learning rate to 3e-5, and the number of training epochs to 4.

\paragraph{Baselines.} We first introduce a naive bert2BERT baseline named DirectCopy, which directly copies the small model to the target model and randomly initializes the unfilled parameters. StackBERT~\citep{stack} and MSLT~\citep{stack2} are also included as the baselines. Both of them are trained in a progressive manner. Following the original setting, for the StackBERT, we first train the 3-layer BERT for 5 epochs, stack it twice into a 6-layer BERT and then train it for 7 epochs. In the final step, we stack the 6-layer model into BERT$_{\rm BASE}$ and further train it with 28 epochs. For MSLT, we first perform 4-stage training. In each stage, we add the top 3 layers of the model already trained to the top of the model and then pre-train the new model by partially updating the top 3 layers. Each stage of the partial training process has 8 epochs. Finally, we further perform 20 full-model training epochs\footnote{We have tried the same setting as the original paper with 8 epoch running but it does not achieve the same loss with BERT$_{\rm BASE}$ (1.511 vs. 1.437). } to achieve the same loss as BERT$_{\rm BASE}$ trained from scratch. The baselines are trained using the same optimizer, training steps and warmup steps as the bert2BERT.

\subsection{Results and Analysis}
We demonstrate the effectiveness of the proposed method on SQuAD and GLUE benchmark and the results are shown in Table~\ref{tab:glue_results}. We also represent the loss curves in Figure~\ref{fig:pretrain_loss_curves_page1} and Appendix A. The results show that: (1) both progressive training methods achieve a saving ratio of more than 10\% and StackBERT is better than MSLT; (2) DirectCopy only saves 13.8\% computational costs, which indicates this naive method of directly copying the trained parameters of the source model to the target model is not effective; (3) our proposed methods, FPI and AKI, achieve better performances than the baselines. Although AKI does not follow the function preserving and has a bigger loss than FPI and DirectCopy at the start of training, AKI achieves a faster convergence rate by using the advanced knowledge; (4) by performing the two-stage pre-training on the target model initialized by AKI, we can save $\bm{45.2\%}$ computational costs. Note that the total parameters of the source model are half of those of the target model (54M vs. 110M). The loss of \MODEL\ in Figure~\ref{fig:pretrain_loss_curves_page1} is high at the stage of sub-model training because it represents the average loss of all sub-models. We also compare the attention patterns of the target models initialized by DirectCopy, FPI and AKI. The attention patterns and their discussions are displayed in Appendix D.

\paragraph{\MODEL\ with smaller source model.} We also evaluate \MODEL\ on one different setting, where the source model $\mathcal{S}$(6, 512) is significantly smaller than the target model (35M vs. 110M). The results are shown in Table~\ref{tab:different_setting} and loss curves are displayed in Appendix B. We observe that DirectCopy achieves no efficiency improvement over the original pre-training, which indicates that the significant size gap between the source and target model greatly reduces the benefit of DirectCopy methods. Compared with DirectCopy, our proposed method reduces the computation cost by 23.9\%, which again demonstrates the effectiveness of \MODEL\. We encourage future work to explore the effect of the size and architecture of the source model on \MODEL.

\begin{table}[ht]
	\centering
	\scalebox{0.80}{
		\begin{tabular}{l|c c| c}
			\toprule
			Model & FLOPs & Ratio &  Avg. \\ 
			~ & ($\times$ ${\rm 1e19}$) & (Cost Saving) & ~ \\
			\midrule 
 \multicolumn{4}{l}{ {\it \MODEL\ }: $\mathcal{S}$(6, 512) $\rightarrow$ $\mathcal{T}$(12, 768)} \\
\midrule
DirectCopy & 7.3  & 0\% &  84.2\\ 
\MODEL\ & 5.6  & 23.9\% & 85.5\\
\bottomrule
\end{tabular}}
\caption{ \MODEL\ on other architectures. Avg means the average score of GLUE datasets.}
\label{tab:different_setting}
\end{table}

\paragraph{Effect of sub-model training epochs.} Our training procedure includes two stages: sub-model training and full-model training. Here, we study the effect of the number of sub-model training epochs by performing \MODEL\ on the different settings of $E_b$=\{0, 5, 10, 20\}. The results are presented in Table~\ref{tab:sub_model_training} and the loss curves are displayed in Appendix C.  We observe that our method achieves the best efficiency when the epoch number is set to 5, while a larger or smaller epoch number will bring a negative impact. We conjecture that a few epochs of sub-model training can establish a stable state for the full-model pre-training to achieve a rapid convergence rate, but with the number of epochs of sub-model training increasing, the knowledge obtained from the source model will be destroyed.


\begin{table}[ht]
	\centering
	\scalebox{0.73}{
		\begin{tabular}{l|c c| c}
			\toprule
			Model & FLOPs& Ratio &  Avg. \\ 
			~ & ($\times$ ${\rm 1e19}$) & (Cost Saving) & ~ \\
			\midrule 
 \multicolumn{4}{l}{ {\it \MODEL\ }: $\mathcal{S}$(12, 512) $\rightarrow$ $\mathcal{T}$(12, 768)} \\ 
\midrule
\MODEL\ ($E_b=0$) & 4.5 & 38.4\% &  86.0 \\ 
\MODEL\ ($E_b=5$) & $\bm{4.0}$ & $\bm{45.2\%}$ &  85.7 \\ 
\MODEL\ ($E_b=10$) & 4.1  & 43.9\% & 85.3 \\
\MODEL\ ($E_b=20$) & 5.4  & 25.4\% & 84.3 \\
\bottomrule
\end{tabular}}
\caption{Effect of sub-model training epochs.}
\label{tab:sub_model_training}
\end{table}

\subsection{Application on GPT}
\paragraph{Datasets.} To demonstrate that our method is generic, following the BERT setting, we also use the Wikipedia and Book Corpus in the GPT-training. For the evaluation, we use the datasets of WikiText-2, PTB, and WikiText103 and evaluate these models under the zero-shot setting without fine-tuning on the training set. 

\paragraph{Implementation Details.} We use the architecture of \{$L$=12, $D$=768\} for the GPT target model, and pre-train it with the learning rate of 1e-4 and training epochs of 20. For \MODEL, we use the source model with an architecture of \{$L$=12, $D$=512\}, initialize the target model with AKI, and pre-train it by the full-model training. 

\begin{figure}[ht]
	\centering
	\includegraphics[width=0.40\textwidth]{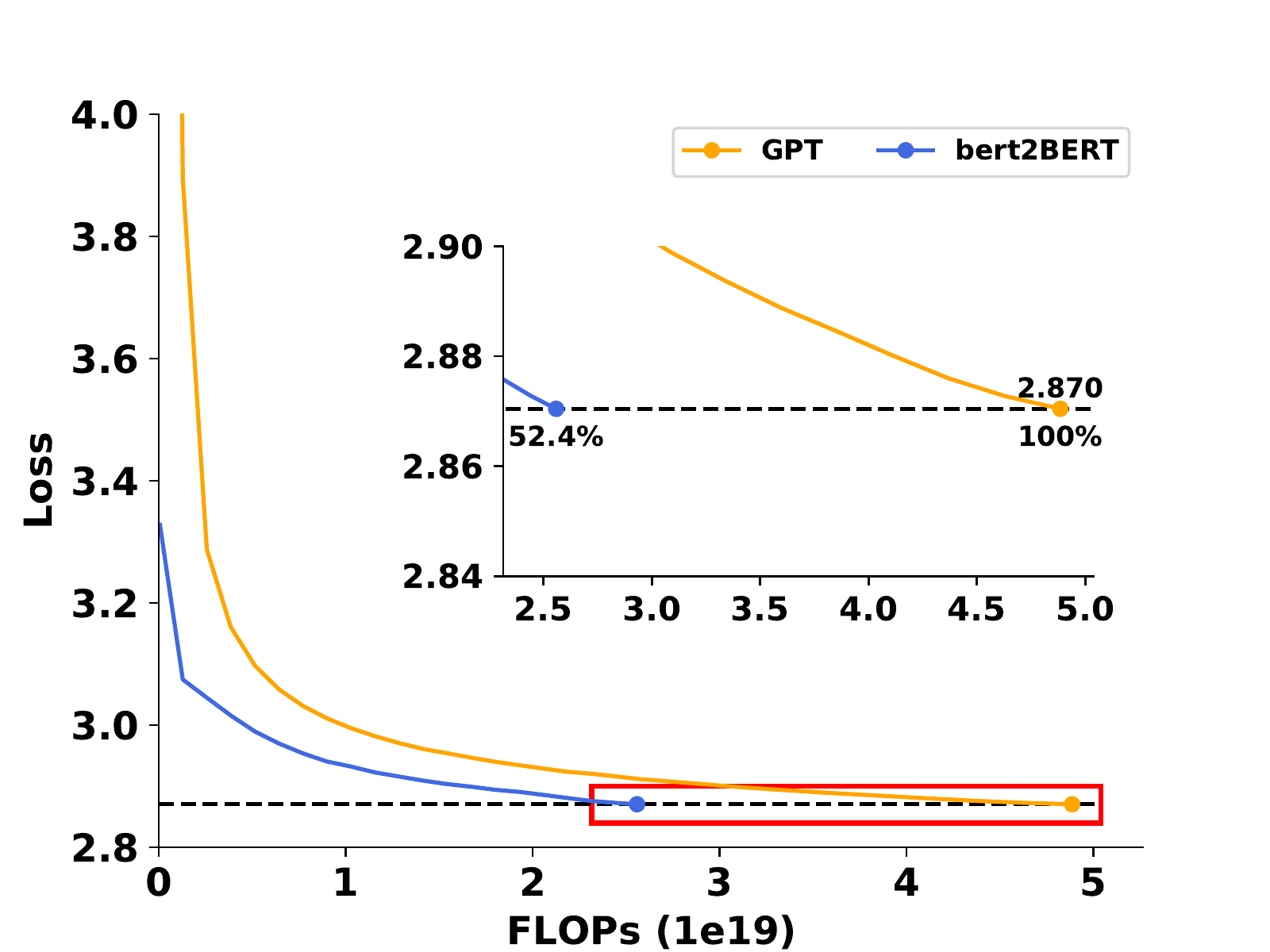}
	\caption{Pre-training loss curves of GPT.}
	\label{fig:gpt_loss_curves}
\end{figure}

\paragraph{Results and Analysis.} We compare the original pre-training method and \MODEL , the results are shown in Table~\ref{tab:gpt_results} and Figure~\ref{fig:gpt_loss_curves}. We observe that the proposed method saves {\bf 47\%} computation cost of GPT pre-training, exhibiting a similar trend to BERT pre-training. Although GPT and BERT have different architectures (e.g., post-LN and pre-LN~\citep{preln}) and are pre-trained with different tasks, \MODEL\ saves a significant amount of training cost on both these two models, which shows that the proposed method is generic and is effective for different kinds of PLMs.

\begin{table}[ht]
	\centering
	\scalebox{0.7}{
		\begin{tabular}{l|c|ccc}
			\toprule
			Model & FLOPs & PTB & WikiText-2  & WikiText103 \\ 
			~ & ($\times$ ${\rm 1e19}$) & (w/o FT) & (w/o FT) & (w/o FT)  \\
			\midrule
			 \multicolumn{4}{l}{ {\it \MODEL\ }: $\mathcal{S}$(12, 512) $\rightarrow$ $\mathcal{T}$(12, 768)} \\ 
			\midrule
            GPT & 4.9 & 133.8 & 47.0  &  53.5  \\
            {\bf \MODEL} & {\bf 2.6} (${\bf 47\%}\downarrow$) & 132.1 &  47.9  & 53.0  \\
            \bottomrule
\end{tabular}}
\caption{Experiments on GPT. We use the perplexity as the metric and ``w/o FT'' means that the pre-trained model is directly evaluated on the test set without fine-tuning on the train set.}
\label{tab:gpt_results}
\end{table} 

\section{Related Work}
{\bf Efficient Pre-training in NLP.} The efficiency in pre-training has been explored by previous work, some try to reuse the parameters of existing PLMs and propose progressive learning. These works~\citep{stack, stack2, growth} are motivated by the fact that different layers have some similar knowledge (e.g., attention patterns). They start pre-training a small model with fewer Transformer layers, and then iteratively expand the model by stacking the already trained layers on the top. Another line of work proposes to “back distil” the knowledge of the small models into large models, which is termed as knowledge inheritance~\citep{inheri}. Another type of work for efficient pre-training focuses on the data efficiency~\citep{wu2021taking} and takes notes for rare words during the pre-training process to help the model understand them when they occur next. The efficiency of pre-training can also be improved by designing efficient pre-training tasks. One classical work is ELECTRA~\citep{electra} which proposes a task of replaced token detection to predict whether each token in the corrupted input was replaced or not. Our method is orthogonal to this kind of work and the combination of ELECTRA and bert2BERT could achieve better efficiency. In addition, there are several other orthogonal techniques for efficient pre-training: mixed precision training~\citep{megatron}, large batch optimization~\citep{lamb}, model architecture innovation~\citep{albert}, layer dropping training technique~\citep{dropping}, etc.

\noindent {\bf Reusable Neural Network.} Reusable neural network, a topic related to transfer learning~\citep{transfer}, is introduced to accelerate the model training in computer vision. One classical work is Net2Net~\citep{net2net}, which first proposes the concept of function-preserving transformations to make neural networks reusable. However, Net2Net randomly selects the neurons to be split. To handle this problem, some works~\citep{splitSteep, escape, fastsplit, Firefly} leverage a functional steepest descent idea to decide the optimal subset of neurons to be split. The pruning technique~\citep{prune} is also introduced for reusable neural networks~\citep{cumulativeTrain}. Recently, hierarchical pre-training is proposed by~\citet{cumulativeTrain}, which saves training time and improves performance by initializing the pretraining process with an existing pre-trained vision model. In this paper, we study the reusable pre-trained language model and propose a new method, \MODEL\, to accelerate the pre-training of BERT and GPT. 
\section{Conclusion and Future Work}
This paper proposes a new efficient pre-training method, \MODEL, which reuses the parameters of the small trained model as the initialization parameters of the large model. We employ the proposed method in both BERT and GPT under different settings of model sizes. The extensive results show that our method is generic to Transformer-based models and saves a significant amount of computation cost. Moreover, the detailed analysis shows that our techniques, function-preserving, advanced knowledge initialization, and two-stage pre-training, are all effective. In future, we will apply \MODEL\ on training super large-scale language models and extends its scope to other PLMs variants such as ELECTRA and BART~\citep{bart}.


\bibliography{anthology,custom}
\bibliographystyle{acl_natbib}

\clearpage
\appendix
\section*{Appendices}
\section{Ablation Study of bert2BERT}
\begin{figure}[h]
\centering
	\includegraphics[width=0.45\textwidth]{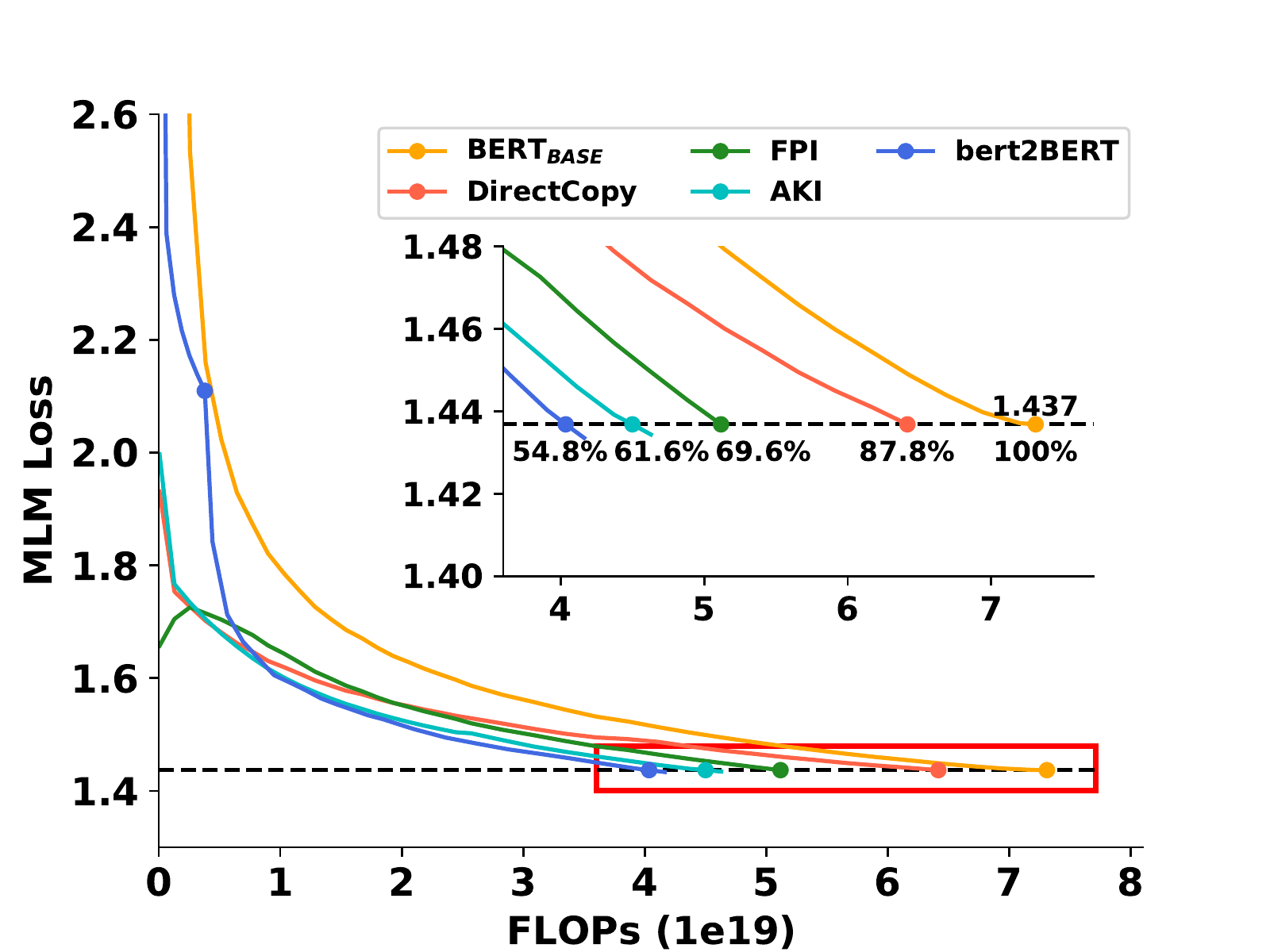}
	\caption{Ablation study of \MODEL . \MODEL\ means the combination of AKI and two-stage pre-training.}
	\label{fig:pretrain_loss_curves}
\end{figure}

We do the ablation study of bert2BERT and the loss curves are displayed in Table~\ref{fig:pretrain_loss_curves}. From the table, we observe that: (1) all the proposed methods is better than the original pre-training method and baseline DirectCopy; (2) although AKI has a worse initialization than FPI, it achieves faster convergence rate than FPI; (3) the two-stage pre-training can further save the total computations from 61.6\% to 54.8\%; (4) the FPI curve has an upward trend at the beginning. We conjecture that it is due to the symmetry brought by FPI and the model needs some optimization time to break this symmetry. 

\section{bert2BERT with smaller source model}
\begin{figure}[h]
	\centering
	\includegraphics[width=0.45\textwidth]{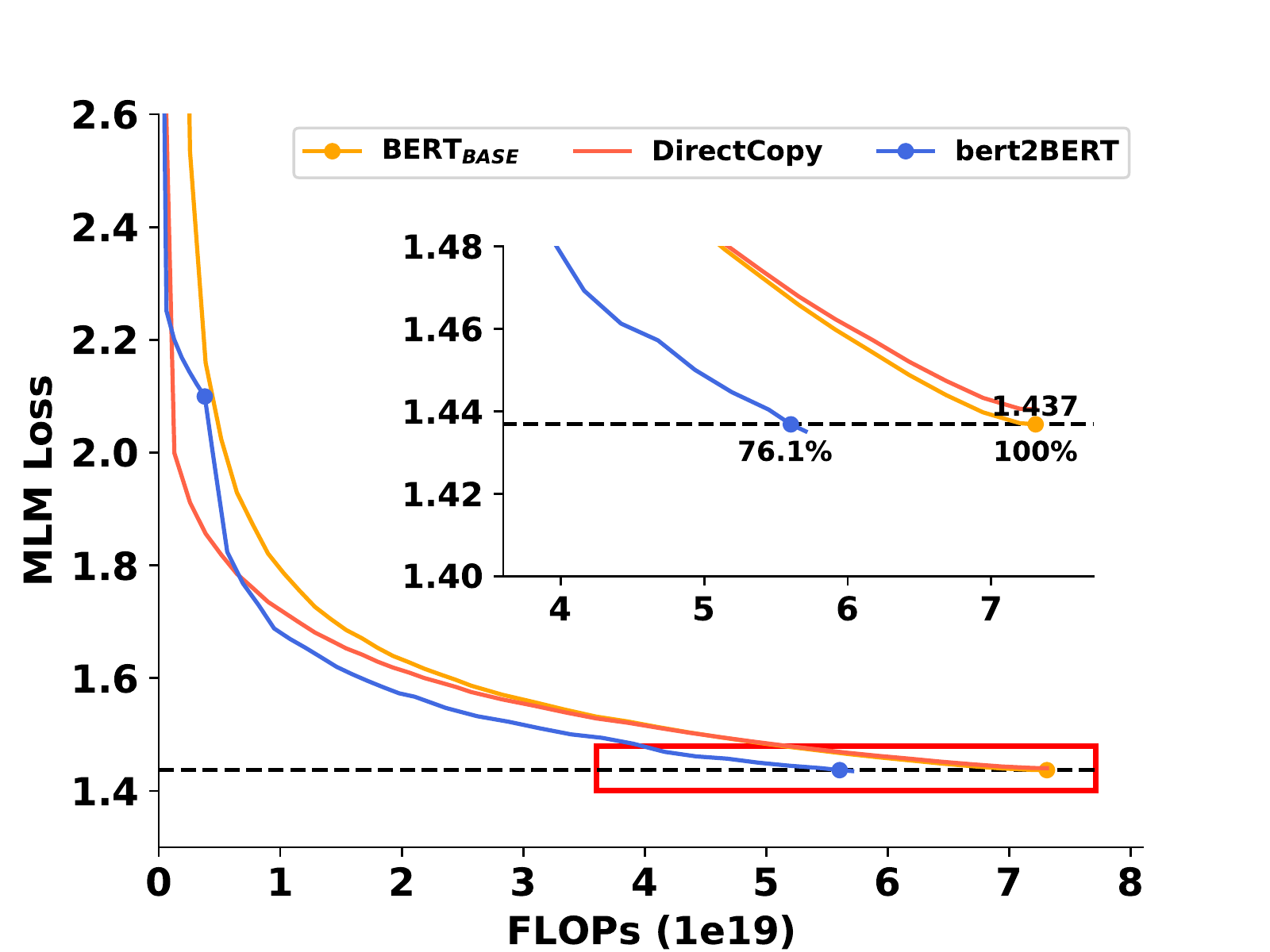}
	\caption{Loss curves of bert2BERT and baselines with smaller source model.}
	\label{fig:loss_curves_smaller_source_model}
\end{figure}

We test \MODEL\ in a different setting with the source model $\mathcal{S}(6, 512)$ and the loss curves are represented in Figure~\ref{fig:loss_curves_smaller_source_model}.

\section{Effect of sub-model training epochs}
\begin{figure}[h]
	\centering
	\includegraphics[width=0.45\textwidth]{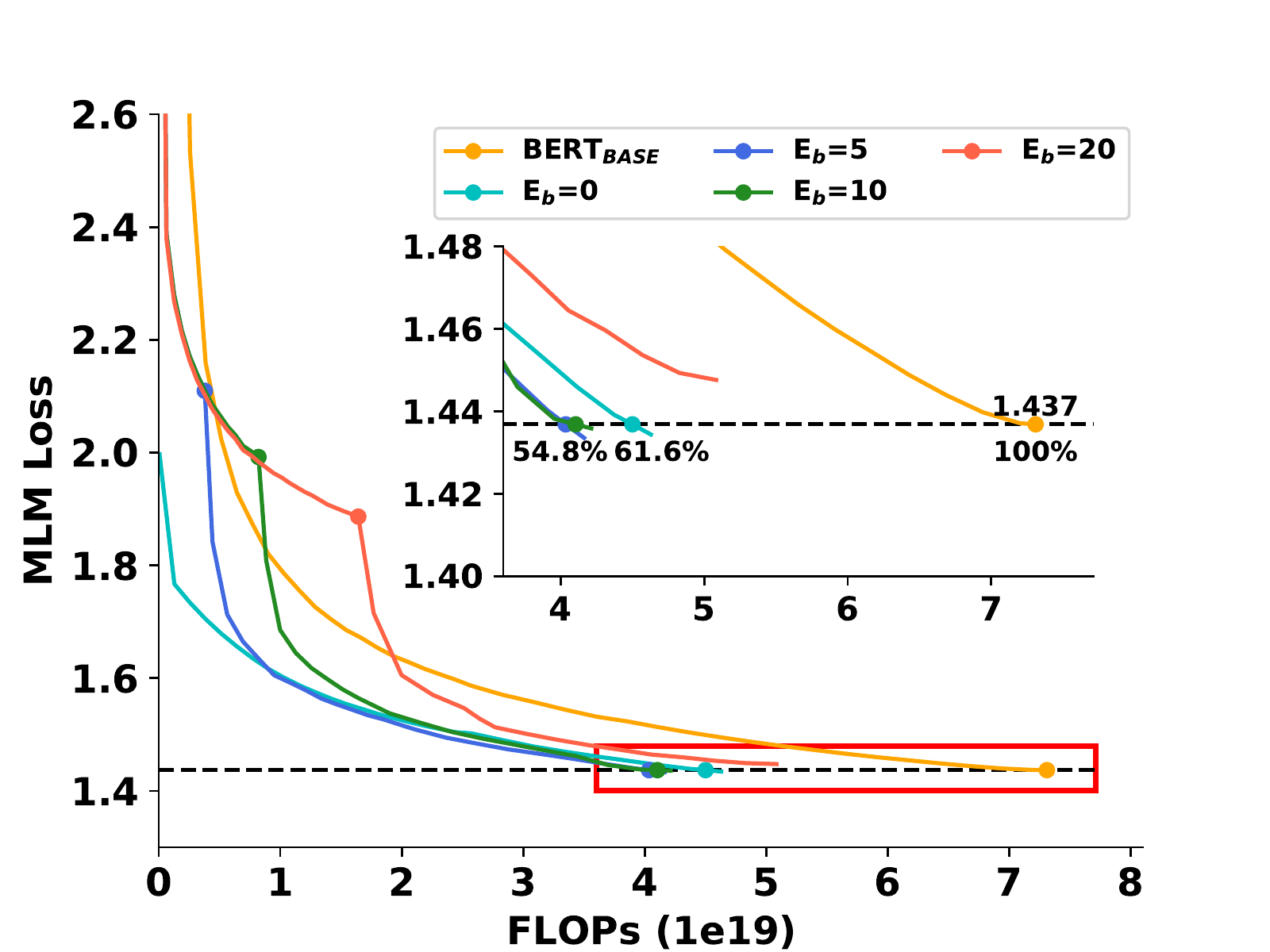}
	\caption{Loss curves of bert2BERT with different sub-model training epochs.}
	\label{fig:loss_curves_different_epochs}
\end{figure} 
We study the effect of sub-model training epochs on the pre-training efficiency. The loss curves are represented in Figure~\ref{fig:loss_curves_different_epochs}. Note that the setting $E_b=20$ has not achieved the same loss (1.437) as the baseline BERT$_{\rm BASE}$ in the 40 training epochs.

\section{Comparisons of Attention Patterns of DirectCopy, FPI and AKI}
\begin{figure*}[t]
\centering
	\includegraphics[width=0.95\textwidth]{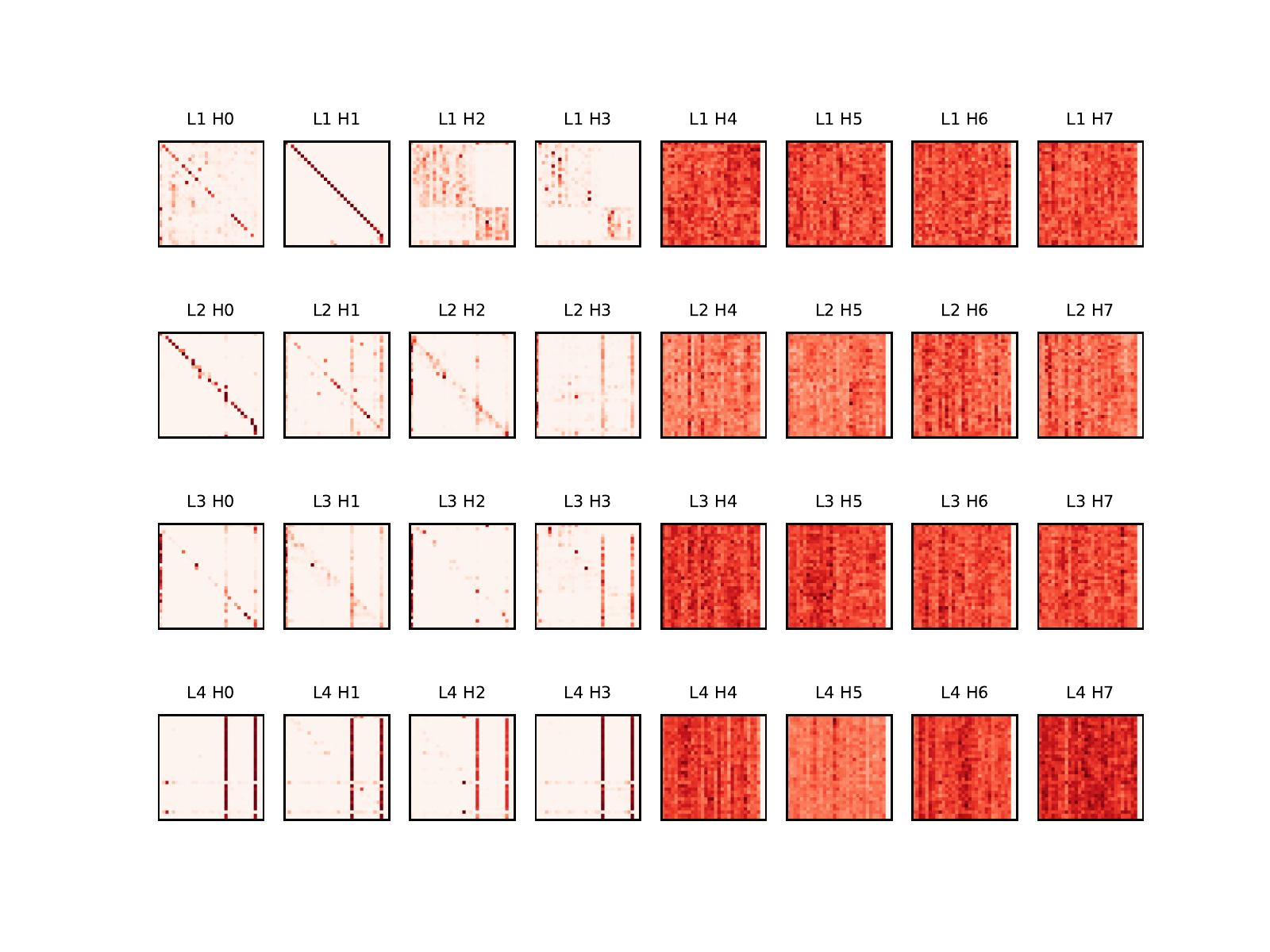}
	\caption{Attention patterns of the target model $\mathcal{T}(4, 512)$ based on the baseline DirectCopy method. The first 4 attention patterns (H0-H3) in each row correspond to the source model's attention patterns, and the last 4 attention patterns (H4-H7) are newly added.}
	\label{fig:attention_modes_l4_h256_dc_l4_h512}
\end{figure*}

\begin{figure*}[t]
\centering
	\includegraphics[width=0.95\textwidth]{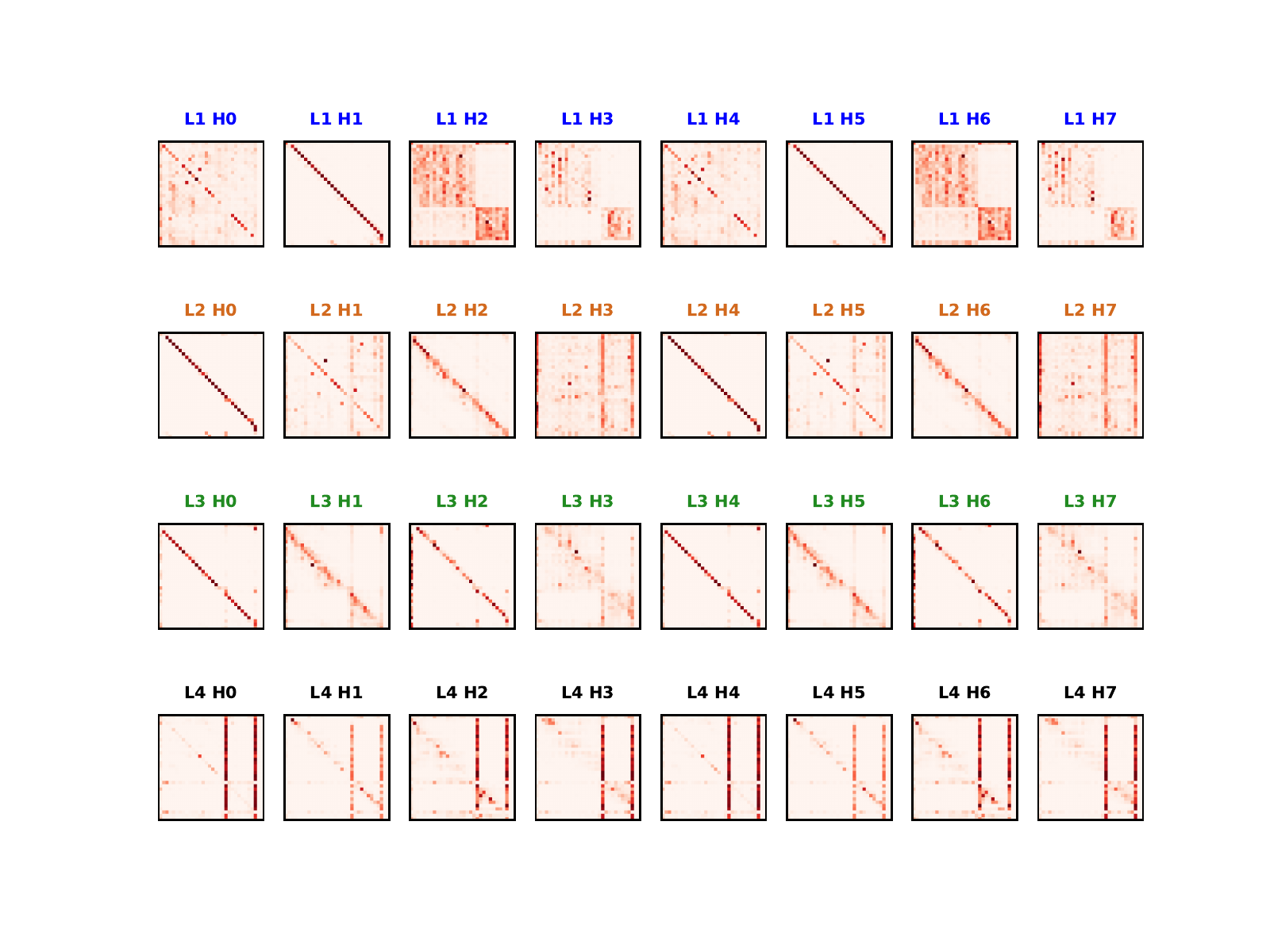}
	\caption{Attention patterns of the target model $\mathcal{T}(4, 512)$ based on our FPI method. The last 4 attention patterns (H4-H7) in each row are obtained by FPI expansion. }
	\label{fig:attention_modes_l4_h256_fpi_l4_h512}
\end{figure*}

\begin{figure*}[b]
\centering
	\includegraphics[width=0.95\textwidth]{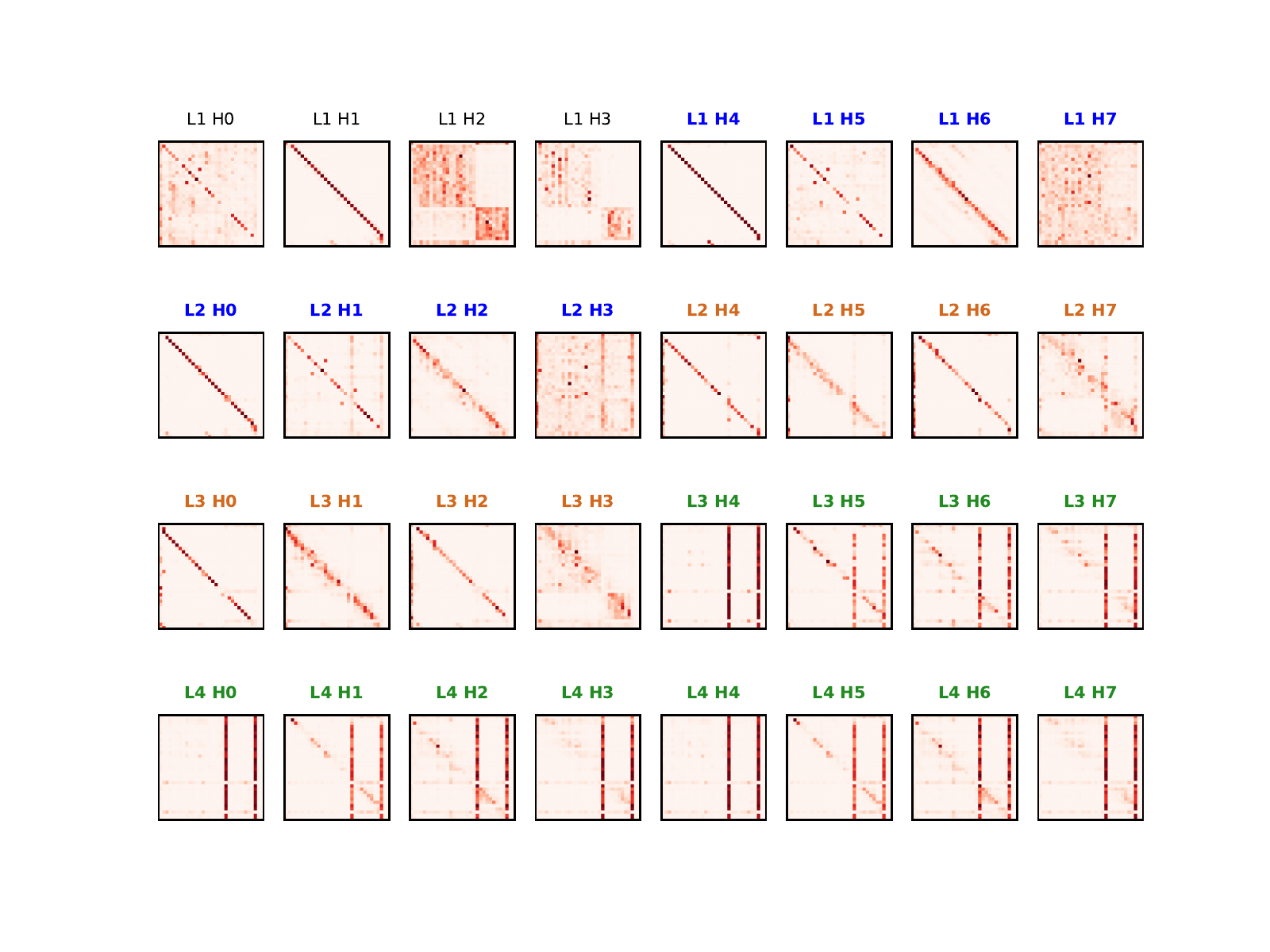}
	\caption{Attention patterns of the target model $\mathcal{T}(4, 512)$ based on our AKI method.  The last 4 attention patterns (H4-H7) in each row are obtained by AKI expansion.}
	\label{fig:attention_modes_l4_h256_aki_l4_h512}
\end{figure*}

We take the source model $\mathcal{S}(4, 256)$ and target model $\mathcal{T}(4, 512)$ as an example to analyze the attention patterns of DirectCopy in Figure~\ref{fig:attention_modes_l4_h256_dc_l4_h512}, FPI in Figure~\ref{fig:attention_modes_l4_h256_fpi_l4_h512} and AKI in Figure~\ref{fig:attention_modes_l4_h256_aki_l4_h512}.

\begin{figure*}[b]
\centering
	\includegraphics[width=0.7\textwidth]{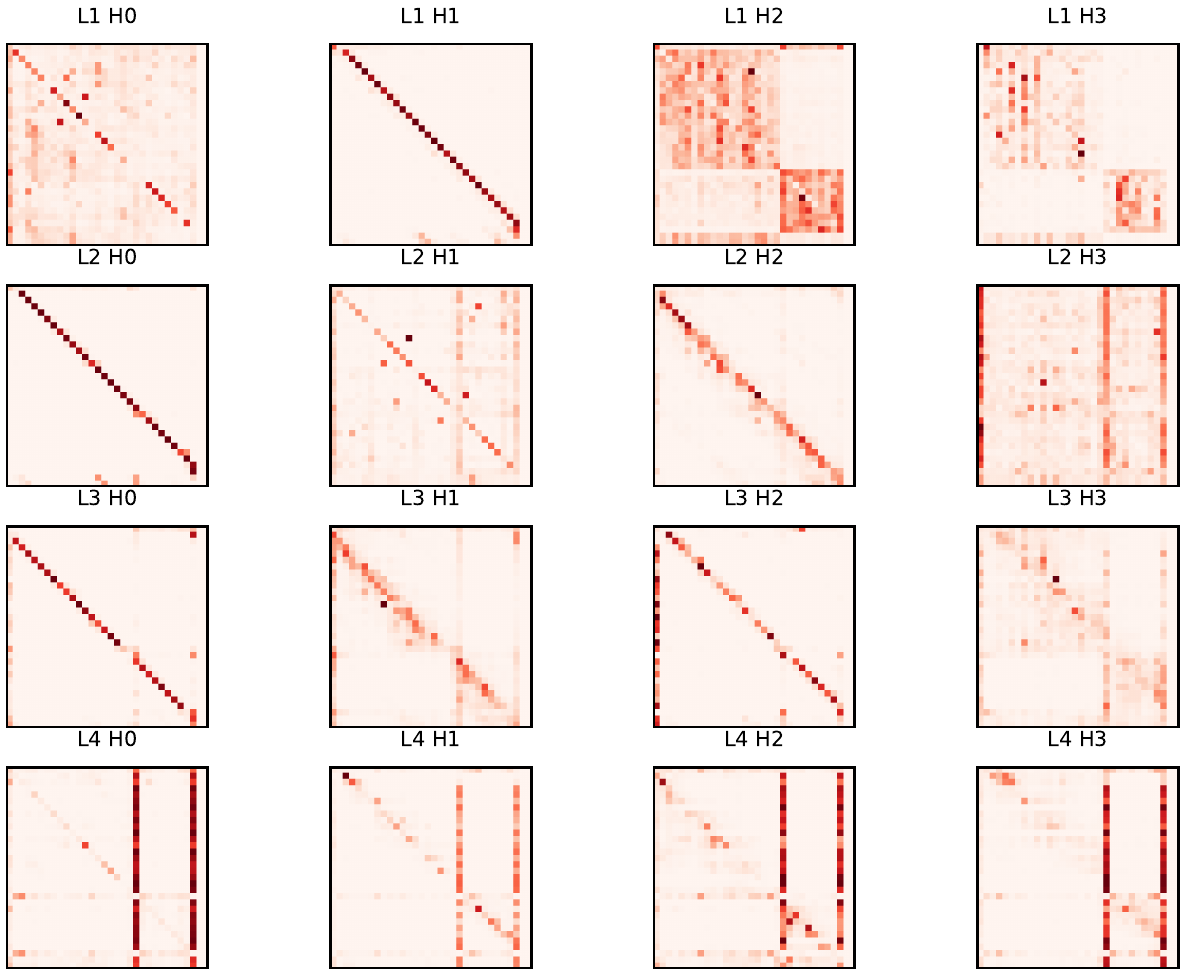}
	\caption{Attention patterns of the source model $\mathcal{S}(4, 256)$, which has 4 attention heads in each layer.}
	\label{fig:attention_modes_l4_h256}
\end{figure*}

We display the attention patterns of the source model $\mathcal{S}(4, 256)$ in Figure~\ref{fig:attention_modes_l4_h256}. Compared with the source model, we observe that the newly added attention patterns of DirectCopy are messy, and the randomly initialized parameters destroy the attention patterns of source model. The proposed FPI method makes new model have the same attention patterns as the source model, thus the knowledge of source model is preserved. However, FPI always induces the symmetrical attention patterns in the same layer. This symmetry will hinder the convergence. To handle this problem, we use AKI method to reuse the parameters of upper layer (advanced knowledge) to break the symmetry, and meanwhile make the knowledge in the same layer richer. Through the AKI method, the attention patterns of upper layer can be also maintained  well in target model. For example, as shown in Figure~\ref{fig:attention_modes_l4_h256_aki_l4_h512}, the attention patterns of 1st layer in target model is similar with the ones of 2nd layer in source model.


\end{document}